\documentclass{article}

\usepackage{arxiv}

\usepackage[utf8]{inputenc} 
\usepackage[T1]{fontenc}    
\usepackage{hyperref}       
\usepackage{url}            
\usepackage{booktabs}       
\usepackage{amsfonts}       
\usepackage{nicefrac}       
\usepackage{microtype}      
\usepackage{lipsum}		
\usepackage{graphicx}
\usepackage{natbib}
\usepackage{doi}
\usepackage[dvipsnames]{xcolor}
\usepackage{tabularx}
\usepackage{array}
\usepackage{amsmath}
\usepackage{cleveref}
\usepackage{multirow}
\usepackage{afterpage}

\usepackage{mathtools}
\usepackage{threeparttable} 
\usepackage{booktabs}
\newcolumntype{Y}{>{\centering\arraybackslash}X}

\usepackage{amssymb}%
\usepackage{pifont}%
\newcommand{\cmark}{\ding{51}}%
\newcommand{\xmark}{\ding{55}}%

\title{RaWaNet: Enriching Graph Neural Network Input via Random Walks on Graphs}

\date{} 					

\author{Anahita Iravanizad\thanks{Chair of Scientific Computing, Department of Mathematics, Technische Universit\"at Chemnitz, 09107 Chemnitz, Germany}
	\And
	Edgar Ivan Sanchez Medina\thanks{Chair for Process Systems Engineering, Otto-von-Guericke University, Universit\"atsplatz 2, Magdeburg, 39106, Germany}
	\And
	Martin Stoll\textsuperscript{*}
}

\hypersetup{
pdftitle={A template for the arxiv style},
pdfsubject={q-bio.NC, q-bio.QM},
pdfauthor={David S.~Hippocampus, Elias D.~Striatum},
pdfkeywords={First keyword, Second keyword, More},
}

\begin{document}
\maketitle

\begin{abstract}

In recent years, graph neural networks (GNNs) have gained increasing popularity and have shown very promising results for data that are represented by graphs. The majority of GNN architectures are designed based on developing new convolutional and/or pooling layers that better extract the hidden and deeper representations of the graphs to be used for different prediction tasks. The inputs to these layers are mainly the three default descriptors of a graph, node features $(X)$, adjacency matrix $(A)$, and edge features $(W)$ (if available). To provide a more enriched input to the network, we propose a random walk data processing of the graphs based on three selected lengths. Namely, (regular) walks of length 1 and 2, and a fractional walk of length $\gamma \in (0,1)$, in order to capture the different local and global dynamics on the graphs. We also calculate the stationary distribution of each random walk, which is then used as a scaling factor for the initial node features ($X$). This way, for each graph, the network receives multiple adjacency matrices along with their individual weighting for the node features. We test our method on various molecular datasets by passing the processed node features to the network in order to perform several classification and regression tasks. Interestingly, our method, not using edge features which are heavily exploited in molecular graph learning, let a shallow network outperform well known deep GNNs.

\end{abstract}

\section{Introduction}

Machine learning based on Euclidean data, such as images, videos or audio signals, has enabled a revolution throughout science and engineering \citep{lecun2015deep}. The backbone of many success stories are deep neural networks and in particular convolutional neural networks. The extension of these techniques to non-Euclidean domains such as social networks, recommender systems, chemistry,  computer graphics, NLP, and etc., has received a tremendous amount of attention \citep{bronstein2017geometric}. In many cases the data are represented as graphs, which are then embedded into graph neural networks (GNNs)\citep{gori2005new, scarselli_graph_2009}. 

The main idea behind GNNs is to incorporate the non-Euclidean data into the network to allow operations such as convolution or pooling. For this the networks need to encode the graph’s adjacency information and possibly initial features for the vertices and/or edges. The two main approaches for GNNs are to design a convolution operator based on either spatial or spectral information of the graph, in order to reveal a hidden and richer representation of the input data. Usually to extract more meaningful features from the graphs multiple spatial/spectral convolutional layers are stack to each other, which makes the network more complex \citep{wu2020comprehensive}.

In order to incorporate the neighborhood information, we propose to process the input data based on random walks of different lengths defined on the graph. The use of random walks is motivated to go beyond a uniform distribution of the nodes and to incorporate a multitude of adjacency information. This will allow us to explore different dynamics within the graph. For this purpose, we calculate the stationary distribution for each walk that scales the node features of the data to serve as further input information to the network. This would equip the network with an enriched initial knowledge, such that we do not always require the use of a deep network or even edge features of the graph. 

We examined our method on various small molecular datasets obtained from MoleculeNet \citep{wu2018moleculenet} and a few molecular combustion-related datasets \citep{schweidtmann2020graph}. Such choice of datasets allows us to test our method on multiple chemical property prediction tasks, and to compare our method with networks that are more complex with respect to the number of layers and the inclusion of bond feature information.

We observe that using the random walk approach combined with a shallow network, not relying on edge features, leads to better or on par performance with the current deep GNN methods for these tasks. Remarkably, this shows just how much information can be used from the graph properties before entering the realm of convolutional or other network layers.

Our results highlight that using enriched initial information allows for simpler and cheaper networks while not sacrificing accuracy. We believe that this technique can lead to new ideas for other challenging tasks in graph-based learning.

\section{Related Works}

Special interest has been set on molecular graphs. The models developed for tasks in this domain are from both categories of classical machine learning and graph neural networks. We believe our method is suitable to be implemented in both realms and to offer prominent advancements in the field of graph neural networks.

\subsection{Graph Neural Networks}
GNNs mainly evolve around designing convolutional layers, which are the generalization of the convolution operation on grid-structured data to the data represented by a graph \citep{wu2020comprehensive}. Graph convolutional operations divide into two categories, spectral-based and spatial-based, which then gave rise to Graph Attention Networks (GANs), Graph Generative Networks (GGNs) and Graph Auto-encoders (GAEs)\citep{cao2020comprehensive}.

\paragraph{Spectral-based Methods.}

The idea behind these methods is to generalize the spectral convolution from Euclidean domain to graph domain, which was initially proposed by  \citep{bruna2013spectral}. The method is to calculate the graph Fourier transform of the input signal, using the spectral decomposition of the graph Laplacian. Due to high computational complexity of eigen-decomposition, several simplifications and approximations of the calculation have been introduced by \citep{defferrard2016convolutional, kipf2016semi, levie2018cayleynets}. Later \citep{zhuang2018dual} introduced Dual Graph Convolutional Network
(DGCN), which is capable of encoding both global and local information of the input data in parallel \citep{wu2020comprehensive}.

\paragraph{Spatial-based Methods.}

As a more straightforward and explicit generalization of CNNs, these methods use the spatial relations of a node to extract its local features with respect to a neighborhood. The idea began with Neural Network for Graphs (NN4G) \citep{micheli2009neural}, which directly adds (sums) up a node’s neighborhood
information at each layer. Similar to the convolution on an image the hidden representation of a node is extracted through aggregation of the information from its neighboring nodes, unlike images the neighborhood of each node varies in size, which is the main challenge to this approach.
Many  different variants of this approach have been proposed to tackle this problem, such as the diffusion convolutional neural network (DCNN) \citep{atwood2016diffusion}, and GraphSAGE \citep{hamilton2017inductive}. Later on \citep{gilmer2017neural} introduced Message Passing Neural Network (MPNN) as a general framework for describing spatial-based convolutional layers.

In order to get more general features, convolutional layers are usually followed
by a pooling layer. Different pooling layers has been designed to learn graph representations, e.g. Set2set \citep{vinyals2015order}, which uses an LSTM-based method and invariant to the order of the elements in the feature matrix $X$, and also is used as a read out function in MPNN to get the graph-level embedding \citep{zhou2020graph}.

The above introduced convolutional and pooling layers are some of the many network layers that are designed to read information from the graph input and are utilized in building up the majority of GNN architectures. The objective of these layers is to learn from the graph's topology and/or its node feature set. The general inputs to graph networks are the set of adjacency information, node features and (possibly) edge features. In this paper, we introduce RaWaNet, a technique that proposes additional inputs to GNN layers. This would enable the network to access further topologies and input feature sets based on different random walks on the graphs. 

\subsection{Molecular Graphs and Conventional Machine Learning}

There are various ways to represent molecular structures, and graph is among the most popular and intuitive ones. In a \textit{molecular graph} vertices and edges are denoting atoms and chemical bonds, respectively. Prior to deep learning approaches in this domain, such as PotentialNet \citep{feinberg2018potentialnet} and MolDQN \citep{zhou2019optimization}, "classical" machine learning approaches were only used to perform property predictions tasks \citep{atz2021geometric}. 

As a prominent set of techniques, Quantitative Structure-Activity Relationship (QSAR) are machine learning models to perform regression and classification tasks for physicochemical, biological and environmental property prediction. The process is to first extract the relevant information contained in the chemical structure of the data and encode it as a set of molecular descriptors, then to construct a model, e.g. \textit{Support Vector Machine (SVM)}, \textit{Decision Tree}, and \textit{Random Forest}, and finally perform various property prediction tasks using the selected molecular descriptors. 2D descriptors/representations (also referred to as topological representations) such as molecular graphs, are among the most popular molecular representations that are effectively used for QSAR modeling \citep{cherkasov2014qsar}.

As a part of our experiments we use the graph embedding produced by our technique to perform a classification task using random forest classifier, where we outperform the results of all the state of the art GNNs and classical models on the HIV molecular dataset \citep{hu2020open} described in the experiments section.

We perform our experiments with the goal to investigate the strength of these additional feature sets via simple and shallow network setting. 

\section{Proposed Method}

Before explaining our method we need to set up some graph notation and preliminaries, and recall graphs properties.

\subsection{Graph Theory Notation and Preliminaries}

A (connected undirected) \textit{graph}, $G=(V,E)$, is defined as a set of \textit{nodes} (vertices) $V=\{v_{1},\ldots, v_{n}\}$ along with a set of \textit{edges} $E=\{\:(v_{i}, v_{j})\:|\:v_{i}, v_{j}\in V \mbox{~and~} i\neq j\} \subseteq V \times V$ whose elements denote connected vertex’s pairs. Cardinally of the node and edge sets are $|V| = n$ and $|E| = m$, respectively. Each two connected vertices are called \textit{adjacent} (or neighbor), written $v_{i}\sim v_{j}$, and that defines the \textit{adjacency matrix} $A$ such that,
\[
  A_{ij} =
  \begin{cases}
    1 & \text{if $(v_{i}, v_{j}) \in E$} \\
    0 & \text{elsewhere.}
  \end{cases}
\]
Assigning a non-negative weight to each edge defines \textit{weighted (undirected) graph}, $G=(V,E,W)$, with the weight matrix $W$ where $W_{ij}=w_{ij} \geq0$. In case of a weighted graph we will have a \textit{weighted adjacency matrix} $A=W$, with $A_{ij}=w_{ij}$.
The \textit{degree} of a node $v_{i}\in V$ is the number of edges adjacent to it and denoted by $d(v_{i})\coloneqq d_{i}=\sum_{j:v_{i}\sim v_{j}}^{} w_{ij}$. The \textit{degree matrix} $D$ is defined as a diagonal matrix whose entries are the degrees of the nodes, i.e. $D=\mathrm{diag}(d_1,\ldots,d_n)$.

A \textit{walk} is an alternating sequence of nodes and edges, $v(0)e_{1}v(1)\ldots e_{k}v(k)$, where $e_{t}=(v(t-1),v(t))$ and $v(t) \in V$ is the walker's position at the (time) step $t$. A \textit{path} is a walk with no repeated vertices, unless it is a closed path, i.e. $v(1)=v(k)$. The \textit{length} of a walk or a path equals to the number of edges in the sequence. The \textit{distance} between two vertices is the length of the shortest path between them. $G$ is \textit{connected} if between each nodes pair there is a path. To find the number of walks of length $k$ from $v_i$ to $v_j$ we calculate $A^k$ and look at the entry of row $i$ and column $j$ (or vice versa, as the undirected graphs have symmetric adjacency matrices).

A \textit{random walk} on a graph $G=(V, E, W)$ is a process of starting from a node $v(0)\in V$ and move to one of its adjacent nodes by random, and repeat the same after landing on the next node until reaching the desired length. This is equivalent to creating a sequence of random nodes, $(v(t),\: t=1,2,\ldots,k)$, where each two consecutive nodes in the sequence are connected. At step $t$, when we are at a node $v(t) = v_{i}$, the probability of arriving at its neighbor node $v_{j}$ is $\Pr (v(t+1)=v_j \:|\:v(t)=v_i) = p_{ij}$ where
\begin{equation}\label{eq:1}
  p_{ij} =
  \begin{cases}
    w_{ij}/d_{i} & \text{if $(v_{i}, v_{j}) \in E$} \\
    0 & \text{elsewhere,}
  \end{cases}
\end{equation}
and forming the \textit{transition matrix} $M=(p_{ij})_{i,j \in V}=D^{-1}W$ \citep{lovasz1993random}. The \textit{probability distribution} of a random node at time (step) $t$ is vector $P_{t}$ where
\begin{equation}\label{eq:2}
  P_{t}(i) = \mathrm{Pr}(v(t) = v_{i}).
\end{equation}
The random walk is then describing a Markov chain (process), with the transition matrix $M$. Therefore the distribution of random walker at step $t$ is
\begin{equation}\label{eq:3}
  P_{t+1} = M^T P_{t}.
\end{equation}
Starting the walk with the distribution vector $P_{0}$ the distribution vector at step $t$ will be
\begin{equation}\label{eq:4}
  P_{t} = (M^T)^t P_{0}.
\end{equation}
A distribution (of a Markov process) is called \textit{stationary} if it won't change in time, i.e. $P_0$ is a \textit{stationary distribution} if $P_t=P_0$ for any $t$. If the distribution of our random walker is stationary then we call our walk a \textit{stationary walk}. In other terms the random walk's stationary distribution is the eigenvector of the transition matrix $M$ corresponding to the eigenvalue $1$, i.e.
\begin{equation}\label{eq:5}
  \pi = M^T \pi.
\end{equation}
As a random walk backwards is also considered a random walk our Markov chain has the \textit{time reversibility} property, which is equivalent to satisfying the \textit{detailed balance} equations
\begin{equation}\label{eq:6}
  \pi_i p_{ij} = \pi_j p_{ji},
\end{equation}
which is easy to solve. We have
\[
  \pi_i \dfrac{w_{ij}}{d_i} = \pi_j \dfrac{w_{ji}}{d_j},
\]
where, as the graph is undirected, $w_{ij}=w_{ji}$. This means $\dfrac{\pi_{i}}{d_i} = \dfrac{\pi_{j}}{d_j}=c$, where $c$ is a constant. Also we know that 
\[
\sum_{i=1}^{n} \pi_{i}=1 \;\:\Rightarrow\;\: c=\dfrac{1}{\sum_{i=1}^{n}d_i},
\]
and hence,
\begin{equation}\label{eq:7}
  \pi = \dfrac{\mathrm{diag}(D)}{\sum_{i=1}^{n}d_i}.
\end{equation}

The \textit{graph Laplacian} of the weighted undirected connected graph $G$ is a symmetric matrix $L=D-A$, with the spectral decomposition $L=U \Lambda U^T$, where $\Lambda = \mathrm{diag}(\lambda_{1},\ldots, \lambda_{n})$.  \citep{riascos_fractional_2014} introduced the \textit{fractional Laplacian} as
\begin{equation}\label{eq:8}
\begin{gathered}
L^{\gamma} = U \Lambda^{\gamma} U^T, \\
\Lambda = \mathrm{diag}(\lambda_{1}^{\gamma}, ..., \lambda_{n}^{\gamma}),\quad \gamma \in (0, 1),
\end{gathered}
\end{equation}
in order to define a random walker who does not tend to revisit the same nodes and therefore explores the network more efficiently. This gives us the adjacency matrix
\begin{equation}\label{eq:9}
  A_\gamma = \mathrm{diag}(L^\gamma) - L^\gamma.
\end{equation}
With the same process as before they have also calculated the stationary distribution of the fractional walk as
\begin{equation}\label{eq:10}
  \pi_\gamma = \dfrac{\mathrm{diag}(L^\gamma)}{\sum_{i=1}^{n}(L^\gamma)_{ii}}.
\end{equation}
The authors have shown that in the fractional random walk, i.e. $0<\gamma<1$, the random walker more often explores and navigates new regions with every step, using long-range
navigation from one node to another arbitrarily distant node and furthermore, the transition probability $M_\gamma=\mathrm{diag}(L^\gamma)^{-1} A_\gamma$ contains a global (non-local) dynamics in the network as apposed to the regular random walks, i.e. $k\in\mathbb{N}$.

\subsection{Random Walk Data Processing: RaWaNet}

As graph data carries such an expressive structure compared to other domains, our suggested model includes more initial information for the input to the graph network than only the matrix of \textit{node features} $X\in \mathbb{R}^{n\times c}$, adjacency information $A$, and if available the \textit{edge features} $X^e\in \mathbb{R}^{m\times d}$.

Random walks of various lengths, as an approach to navigate on the network, each construct a new adjacency matrix with the weights $A_{ij}^k = w_{ij}^{(k)}$ being the number of possible walks of length $k$ between the two nodes, or in other words a measure of the strength of their connection.

For the lengths $k\in \{1,2\}$ the walk is also a path between nodes pairs. As we know, if the walks are of the regular length $k=1$ there would be no repeated nodes in every walk between two nodes. The same holds for the walks of length $k=2$, except for the loops, i.e. the diagonal entries of $A^2$, which are closed paths. Since the loops do not help with the navigation on the graph, and do not add to the information of the nodes (as we already used the neighborhood of the nodes in $A_1$) we remove the loops from the adjacency matrix.

This gave rise to our idea of creating multiple further sets of adjacency matrices for the input graphs and learn the data with more enriched initial information. We calculate the stationary distributions of each random walk to use as a weighting for the node features. 

\begin{figure*}[ht!]
  \centering
  \includegraphics[width=\linewidth]{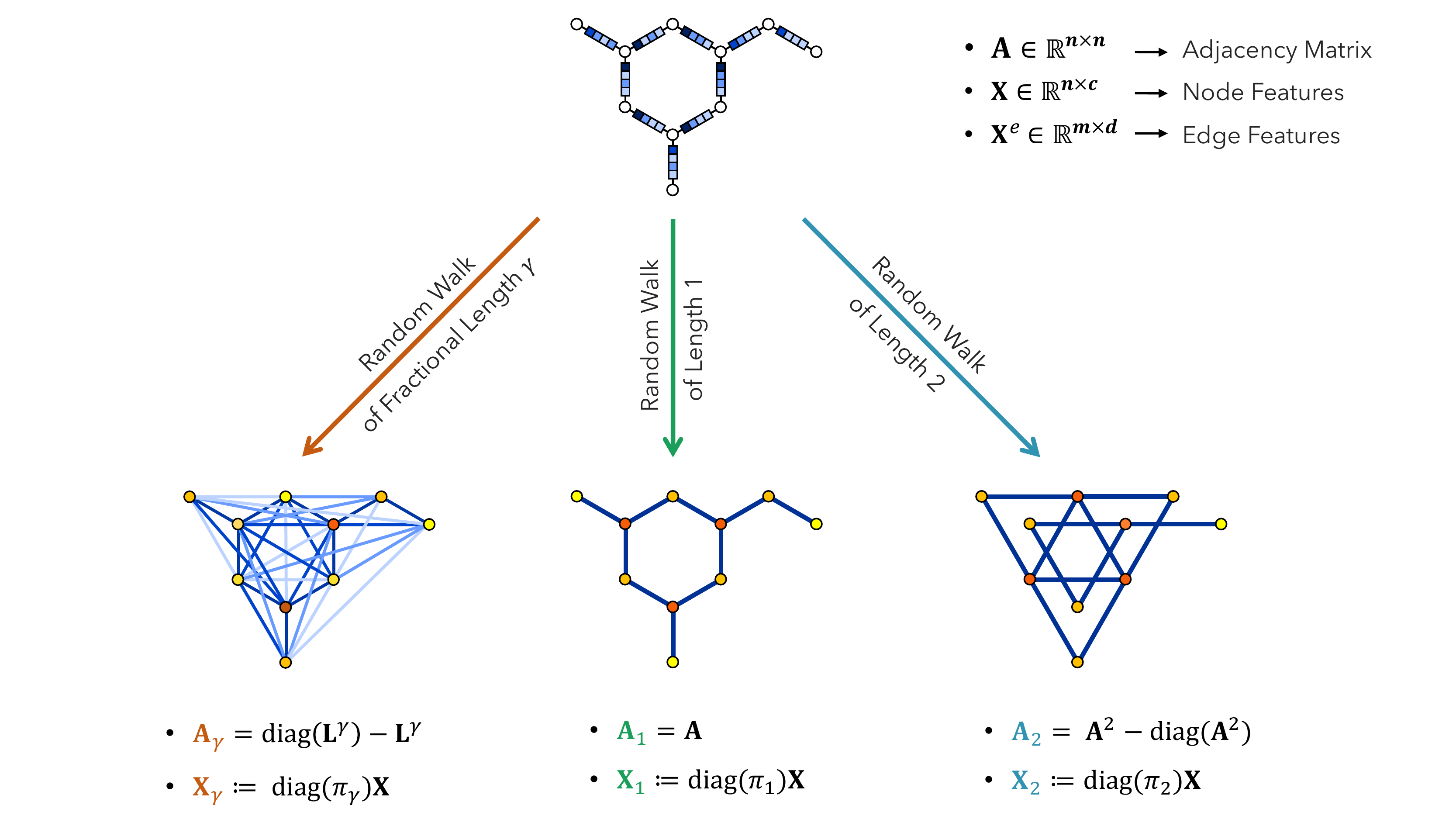}
  \caption{Illustration of an input graph with its three proposed random walks. $A_1$ and $A_2$ for expressing the local dynamics and $A_{\gamma}$, with $0<\gamma<1$, for a global and more descriptive navigation of the network. The entries of the adjacency matrix of each graph equals the number of walks between each two nodes, which is shown via thickness and color of the edges. $\mathrm{diag}(\pi)$ denotes a diagonal matrix, where the diagonal entries are the vector $\pi$, the stationary distribution of the random walker. The colored nodes indicate the weighted node features obtained by multiplication of the stationary distribution of each random walk with the node features.
}
  \label{fig:rw}
\end{figure*}

To construct both local and non-local dynamics of the graph we selected two standard random walks, i.e. of the length $k\in \{1,2\}$, and a fractional random walk of length  $0<\gamma<1$, respectively. In addition to $(A, X, X^e)$ our input to the graph (neural) network contains the following information:
\begin{align}
    (A_1, X_1), \label{eq:11} \\
    (A_2, X_2), \label{eq:12} \\
    (A_\gamma, X_\gamma), \label{eq:13}
\end{align}

where 
\begin{equation*}
\begin{aligned}
A_1 &= A, \hspace{7.5em} X_1 = \mathrm{diag}{(\pi_1)}X, \\
A_2 &= A^2 - \mathrm{diag}{A^2}, \hspace{2.8em}
X_2 = \mathrm{diag}{(\pi_2)}X, \\
A_\gamma &= \mathrm{diag}(L^\gamma) - L^\gamma, \qquad
X_\gamma = \mathrm{diag}{(\pi_\gamma)}X. \\
\end{aligned}
\end{equation*}

We know that the stationary distribution vector $\pi$ is the eigenvector of the transition matrix $M$. Therefore,
\begin{equation*}
\mathrm{diag}{(\pi)} \in \mathbb{R}^{n \times n},
\end{equation*}
\begin{equation*}
\mathrm{diag}{(\pi)}X \in \mathbb{R}^{n \times c}.
\end{equation*}
In Figure~\ref{fig:rw} we give an illustration of the proposed random walks with their adjacency matrices and modified node features. The proposed additional set of graph inputs in \Cref{eq:11,eq:12,eq:13} can then be used as input to proposed GNN layers, such as convolutional, pooling, normalization, etc.

There are various ways to select and make use of the additional inputs in order to enhance learning the data. From using all of them along with the initial graph inputs, to only using one, e.g. $(A_2, X_2)$. This could give rise to further research on how for different datasets and different network architectures a different subset of these inputs improve the learning accuracy.

\subsection{A Solution for Disconnected Graphs}

For this method we need the graphs to be connected, while we know some data can include disconnected graphs. For example in our experiments, the molecular graphs might contain ions which causes disconnection in graphs.

To address this problem we tracked the disconnected graph and added an extra node to it, whose features are zero and is connected to all the nodes in the graph. In other words, a zero-padding for $X$ from below, and a one-padding for $A$ from below and right (except for the diagonal entry). This way the network could yet distinguish graphs that are originally disconnected from the connected ones. 

We also need $A^2$ to be a connected graph, therefore the graphs need to have at least three vertices. For the graphs with two vertices we followed the above, and fixed the problem with an additional node. There are rare exceptions of having graphs with a single node due to the approach of constructing hydrogen depleted graphs, e.g. methane. For these cases we proposed adding two additional nodes that are connected to the single node and to each other. This leads to add two rows of zeros to $X$ and two rows and columns of ones to $A$ (except for the diagonal entries).

There are certainly other approaches to prepare the dataset to be connected with minimum three nodes, which could be further investigated.

\section{Experiments}\label{sec:4}

The recent IPCC report \citep{IPCC_2021} has highlighted, once more, the urgent need for a transition to a more sustainable economy. In this regard, the enormous chemical space has to be explored to find potential sustainable replacements of many currently used chemical compounds. For encouraging the contributions towards solving this worldwide-concerning conundrum, we have used three molecular combustion-related datasets (MON, RON, DCN) \citep{schweidtmann2020graph} that reflect the autoignition behavior of molecules. This might lead to substantial advances in the design/discovery of renewable fuels. Additionally, we have applied our method to a very small ecotoxicological dataset (BCF)\citep{zhao2008new}, that meets the official REACH legislation \citep{REACH}, in a way to test our model with very small molecular graph databases (a common situation in most sustainability related chemical data banks \citep{VEGA_QSAR}). This dataset reports the bioacummulation factor of several chemicals in fish, which is closely related to the accumulation of pollutants in ecosystems. Moreover, in order to ensure a fair comparison to other existing learning algorithms, we have also tested our method using several molecular datasets available from MoleculeNet \citep{wu2018moleculenet} and splitted by OGB \citep{hu2020open}. This includes biophysical (HIV, BACE), physico-chemical (ESOL, FreeSolv, Lipophilicity) and physiological (BBBP, SIDER, ClinTox) properties.

Many graph networks have been developed in recent years and their performance has been evaluated for learning molecular datasets. To investigate the informativity of the proposed additional inputs, i.e. \Cref{eq:11,eq:12,eq:13}, we perform our experiments on a shallow network where its configuration is tuned for each dataset separately. For a more precise comparison with the results of state of the art methods, we choose the same setting for atom features. The aim of our experiments is not to outperform the current results, but to highlight  the information contained in the data processing we propose. As a result we believe that this can  be implemented in various current and future graph network architectures.

Current state of the art methods working on molecular graphs are based on either neural fingerprints \footnote{Here by neural fingerprints we refer to the final graph embedding vector obtained from the GNN models.} that combined atom and bond features in an optimized way according to the task at hand or circular fingerprints (e.g ECFP) that enconde the substructures present in the molecular graph into a fixed-size vector \citep{duvenaud2015convolutional}.  For a fair comparison, we use the same data splitting and SMILES string transformation to molecular graphs as the networks we are comparing our results to. We also keep the same transformation of the initial features $X$ to a higher dimension for each task as is done in deep methods explained later.

The extended input to all the networks for our experiments is
\[(A, X, X^{e}, A_1, X_1, A_2, X_2, A_\gamma, X_\gamma).
\]
For the fractional walk length we chose $\gamma = 0.1$. Also, we did not use the bond features $X^e$ in our experiments because we would like to highlight the sole contributions from the additional weighted features that results from our random walk technique.

\subsection{OGB Classification and Regression Tasks}

\begin{table*}[ht!]
\begin{center}
\begin{threeparttable}[ht!]
\centering
\caption{ROC-AUC results for OGB binary classification datasets. Here the shallow network is a one prediction (linear) layer. The comparison is with the performance of various models, taken from OGB leaderbord.}
\label{tab:Class}
\begin{tabularx}{\linewidth}{llXXYY}
\noalign{\hrule height 1pt}
\multicolumn{1}{p{1cm}}{\centering \textbf{Dataset}} & 
\multicolumn{1}{p{1cm}}{\centering \textbf{Model}} & 
\multicolumn{1}{p{1cm}}{\centering \textbf{Test}} & 
\multicolumn{1}{p{1cm}}{\centering \textbf{Valid.}} & 
\multicolumn{1}{X}{\centering \textbf{Edge Feat.}} & \multicolumn{1}{X}{\centering \textbf{\#Tasks}}\\
\noalign{\hrule height 1pt}
\multicolumn{1}{l}{\multirow{9}{*}{\normal{\texttt{ogbg-molhiv}}}} & {\textbf{FingerPrint + GMAN}} & \textbf{82.44\%} & 83.29\% & {\cmark} & \multirow{9}{*}{1}\\
& {Neural FingerPrints} & 82.32\% & 83.31\%  & {\cmark} \\
& Graphormer + FPs & 82.25\% & 83,96\% & {\cmark}\\
& Graphormer \citeyearpar{ying2021transformers} & 80.51\% & 83.10\% & {\cmark} \\
& directional GSN \citeyearpar{bouritsas2020improving} & 80.39\% & 84.73\% & {\cmark}\\
& RaWaNet{\small{ + Shallow Network}} & 79.79\% & 81.12\% & {\xmark} \\
& DGN \citeyearpar{beani2021directional} & 79.70\% & 84.70\% & {\cmark}\\
& {EGC-M} \citeyearpar{tailor2021adaptive} & 78.18\% & 83.96\% & {\xmark}\\
& {GIN + virtual node} \citeyearpar{hu2020open} & 77.07\% &  \textbf{84.79\%} & {\cmark}\\
\noalign{\hrule height 0.5pt}
\multirow{2}{*}{\normal{\texttt{ogbg-molsider}}} 
& \textbf{RaWaNet{\small{ + Shallow Network}}} & \textbf{62.58\%} & \textbf{61.22\%} & {\xmark}  & \multirow{2}{*}{27} \\
& {GCN + virtual node} \citeyearpar{hu2020open}& {61.65\%} & {59.86\%}  & {\cmark} & \\
\noalign{\hrule height 0.5pt}
\multirow{2}{*}{\normal{\texttt{ogbg-molclintox}\quad}} & \textbf{RaWaNet{\small{ + Shallow Network}}}& \textbf{92.01\%} & {96.95\%}  & {\xmark} & \multirow{2}{*}{2}\\
& {GCN} & {91.30\%} & \textbf{99.24\%} & {\cmark} & \\
\noalign{\hrule height 0.5pt}
\multirow{2}{*}{\normal{\texttt{ogbg-molbbbp}}}
& \textbf{RaWaNet{\small{ + Shallow Network}}} & \textbf{70.27\%} & \textbf{96.43\%} & {\xmark}  & \multirow{2}{*}{1}\\
& {GIN + virtual node} & {69.88\%} & {94.66\%} & {\cmark} & \\
\noalign{\hrule height 0.5pt}
\multirow{2}{*}{\normal{\texttt{ogbg-molbase}}}
& \textbf{RaWaNet{\small{ + Shallow Network}}} & \textbf{83.37\%} & \textbf{76.83\%} & {\xmark} & \multirow{2}{*}{1}\\
& {GCN} & {79.15\%} & {73.74\%} & {\cmark} & \\
\noalign{\hrule height 1pt}
\end{tabularx}
\end{threeparttable}
\end{center}
\vskip -0.01in
\end{table*}

\begin{table*}[ht!]
\begin{center}
\begin{threeparttable}[ht!]
\centering
\caption{RMSE results for OGB regression tasks. Here the shallow network is a one prediction (linear) layer.}
\label{tab:Reg}
\begin{tabularx}{\linewidth}{llYYY}
\noalign{\hrule height 1pt}
\multicolumn{1}{l}{\textbf{Dataset}} & \multicolumn{1}{l}{\textbf{Model}} & \multicolumn{1}{c}{\textbf{Test}} & \multicolumn{1}{c}{\textbf{Valid.}} & \multicolumn{1}{c}{\textbf{Edge Feat.}} \\
\noalign{\hrule height 1pt}
\multirow{2}{*}{\normal{\texttt{ogbg-molesol}}} & \textbf{RaWaNet{\small{ + Shallow Network}}} & \textbf{0.922} & 0.882 & {\xmark} \\
& {GIN + virtual node} & {0.998} & \textbf{0.878} & {\cmark} \\
\noalign{\hrule height 0.5pt}
\multirow{2}{*}{\normal{\texttt{ogbg-molfreesolv}}} & \textbf{RaWaNet{\small{ + Shallow Network}}} & \textbf{1.905} & {2.282} & {\xmark} \\
& {GIN + virtual nod} & {2.151} & \textbf{2.181} & {\cmark} \\
\noalign{\hrule height 0.5pt}
\multirow{2}{*}{\normal{\texttt{ogbg-mollipophilicity\quad}}} & \textbf{GIN + virtual node} & \textbf{0.704} & \textbf{0.679} & {\cmark} \\
& RaWaNet{\small{ + Shallow Network}} & 0.813 & 0.800 & {\xmark} \\
\noalign{\hrule height 1pt}
\end{tabularx}
\end{threeparttable}
\end{center}
\vskip -0.01in
\end{table*}

OGB \citep{hu2020open} has adopted these datasets (HIV, BACE, ESOL, FreeSolv, Lipophilicity, BBBP, SIDER, ClinTox)  from MoleculeNet and performed a standardized scaffold splitting procedure on them. As we mentioned before, the network setting for our experiments is shallow and has a simple setup. For all datasets we created an initial node embedding using the OGB atom-encoder class. The dimension of the embedding is tuned for each dataset and is set to 1000 for the \texttt{ogbg-molhiv} dataset and is either 150 or 300 for all other datasets. A subset of $\{X, X_1, X_2, X_\gamma \}$ is selected for each task, where $\gamma=0.1$. Each task is using 3 or 4 of theses feature sets except for the \texttt{ogbg-molhiv} dataset, where we only use $X_2$ to learn. For all datasets but \texttt{ogbg-molhiv} we apply a graph normalization, GraphNorm \citep{cai2021graphnorm}, on each of the feature matrices, followed by an activation function of either ReLU, tanh, or sigmoid. The results are then passed through a global pooling layer, which is either the average pooling alone or the average pooling scaled via max pooling. The output vectors are then concatenated and fed into a linear layer to perform the according prediction tasks. We use the Adam optimizer for all the methods and tune the learning rate for each task. The batch size for train/validation/test is 32 across all datasets. Regression tasks are trained for 300 epochs, \texttt{ogbg-molhiv} for 50 epochs and all other classification tasks for 100 epochs. We repeat all the experiments 10 times and report the mean of their respective accuracy measure.

The results for the classification tasks are shown in Table \ref{tab:Class}. We can see a higher performance from our learning method compared to the deep graph neural networks. This highlights once more that our network achieves comparable performance while being shallow and significantly simpler.
For the \texttt{ogbg-molhiv} dataset, we include the results by some of the selected models from the OGB leaderbord, including the current top leading models. These provide a variety of baselines with the models ranging from traditional MPNNs to the most recent more expressive architectures.
We observe that the top 3 models with a significant jump in their performance are benefiting from combining two shared factors, the complexity of their models and the use of the extended-connectivity fingerprints (ECFPs). For instance, Graphormer \citep{ying2021transformers} learns over 47M network parameters compared to the number of parameters of our model that is at least 2 orders of magnitude less. These results suggest that a meaningful processing of the input data can contribute significantly to the performance of the model. We believe that in the future this can be considered and implemented in a more general way through our RaWaNet framework along different graph domains.

Regression results are shown in Table \ref{tab:Reg} where we outperform the state of the art for two out of three tasks via a shallow network. 

\subsubsection{Classical Machine Learning Approach for \texttt{ogbg-molhiv}}

As mentioned above, the top results in the leaderboard (shown in \Cref{tab:Class}) belong to deep networks trained using molecular fingerprints, more precicely ECFPs, which are vector embeddings encoding the structure of the molecules.

We test our method in a different way and produce a fingerprint for graphs in \texttt{ogbg-molhiv} dataset in order to fit a classical machine learning classifier. This time we extract a one-hot encoding of the same node features as above and use global pooling layers to get the vector embedding.

For this we employ two feature sets, $\{X, X_2\}$, and use the embedding vector to fit a random forest classifier. We use max and average global pooling to output the final feature vector, which then serves as the input to fit the classifier.

We perform the classification task on two models; the first one only takes our RaWaNet feature vector as input while the second one uses the concatenation of both the RaWaNet vector and the the Morgan finger print (\citep{rogers2010extended}). The size of the feature vector in our first model is 513, and in the second one $(2048+513)$. 

The number of the estimators for all models is 1000 and the number of jobs is set to -1. The min\_samples\_leaf in our both models is set to 4. The max\_depth when we only have RaWaNet input is set to 10, while it is set to None for the other models. We show the averaged results over 10 runs.

\begin{table*}[h]
\begin{center}
\begin{small}
\begin{threeparttable}[h]
\centering
\caption{ROC-AUC results for \texttt{ogbg-molhiv} classification task using classical machine learning approach. The second result is taken from OGB leaderbord.}
\label{tab:RF}
\begin{tabularx}{0.7\linewidth}{lcc}
\noalign{\hrule height 1pt}
\multicolumn{1}{X}{\textbf{Model}} & \multicolumn{1}{Y}{\textbf{Test (\%)}} & \multicolumn{1}{Y}{\textbf{Valid. (\%)}} \\
\noalign{\hrule height 1pt}
\textbf{RaWaNet + Morgan FP + Rand. Forest} & \textbf{82.84$\pm$0.22} & 83.17$\pm$0.17 \\
{Morgan FP + Rand. Forest} \citeyearpar{rogers2010extended} & 80.60$\pm$0.10 & \textbf{84.20$\pm$0.30} \\
{RaWaNet + Rand. Forest} & 80.27$\pm$0.13 & 77.33$\pm$0.36 \\
\noalign{\hrule height 1pt}
\end{tabularx}
\end{threeparttable}
\end{small}
\end{center}
\vskip -0.01in
\end{table*}

Our results (\Cref{tab:RF}) indicate how expressive is the embedding produced by our processed data. With a much less deep tree we got on par test accuracy as when we use the Morgan FP. Also the feature vector from RaWaNet is much smaller in size, which can be the result of a more compressed feature extraction in this technique. Finally, combining both vectors results in outperforming all the existing state of the arts (shown in \Cref{tab:Class,tab:RF}), while maintaining a much simpler structure for the network.

\subsubsection{Going Deep for Lipophilicity Dataset.}

To test if going deep with our additional input can improve the performance we implemented a convolutional message passing network for \texttt{ogbg-mollipophilicity} dataset. As we did not perform well learning this dataset with our simple setting, we use a deep network with our proposed RaWaNet inputs. Here, we again do not use the bond feature information. After encoding the initial input node features to an embedding of dimension 50, we concatenate $X_1$, $X_2$, and $X_\gamma$ to obtain one feature tensor $X_T$.  Then $X_T$ is fed into three parallel message passing layers GraphConv \citep{morris2019weisfeiler}  with adjacency information corresponding to walks of  length 1, 2, and $0.1$, respectively. We then pass the outputs through a shared GRU layer to form a message passing network scheme. The resulting features are now the updated $X_1$, $X_2$, and $X_\gamma$, which are again concatenated and passed through GraphConv layers followed by another GRU layer as before. Finally, we apply an average pooling to each updated embedding and sum these three pooled features to then perform a prediction using a linear layer. We use the Adam optimizer with a 0.001 learning rate and run the network for 150 epochs. As it is shown in Table \ref{tab:deepmollip}, we outperform the result from OGB.

\begin{table}[ht!]
\begin{center}
\begin{threeparttable}
\centering
\caption{RMSE results for \texttt{ogbg-mollipophilicity}, learned by a deep network setting with our random walk processed input. Here the Deep Network is a message passing scheme using GraphConv + GRU layers.}
\label{tab:deepmollip}
\begin{tabularx}{0.7\columnwidth}{lccc}
\noalign{\hrule height 1pt} \multicolumn{1}{X}{\textbf{Model}} & \multicolumn{1}{Y}{\textbf{Test}} & \multicolumn{1}{Y}{\textbf{Valid.}} & \multicolumn{1}{Y}{\textbf{Edge Feat.}} \\
\noalign{\hrule height 0.5pt}
\textbf{RaWaNet{{\small{ + Deep Network}}}} & \textbf{0.684} & 0.695
 & {\xmark} \\
GIN + virtual node & 0.704 & \textbf{0.679} & {\cmark} \\
\noalign{\hrule height 1pt}
\end{tabularx}
\end{threeparttable}%
\end{center}
\vskip -0.01in
\end{table}
\begin{table*}[ht!]
\begin{center}
\begin{threeparttable}[ht!]
\centering
\caption{MAE results for RON, MON, DCN  and BCF property prediction tasks. Here the shallow network contains two hidden (linear) layers, for generating feature embedding and prediction. {GNN$_1$} and {GNN$_2$} are from \citep{schweidtmann2020graph} and \citep{medina2021prediction} respectively.}
\label{tab:Reg2}
\begin{tabularx}{\linewidth}{llYYYY}
\noalign{\hrule height 1pt}
\multicolumn{1}{l}{\textbf{Dataset}} & \multicolumn{1}{l}{\textbf{Model}} & \multicolumn{1}{c}{\textbf{Avg. MAE}} & \multicolumn{1}{c}{\textbf{EL MAE}} & \multicolumn{1}{c}{\textbf{EL $\mathbf{R^2}$}} & \multicolumn{1}{c}{\textbf{Edge Feat.}}\\
\noalign{\hrule height 1pt}
\multirow{2}{*}{MON} & \textbf{RaWaNet{\small{ + Shallow Network}}} & \textbf{5.2} & 4.6 & 90\% & {\xmark} \\
& {GNN$_1$} & 5.4 & - & - & {\cmark} \\
\noalign{\hrule height 0.5pt}
\multirow{2}{*}{RON} & RaWaNet{\small{ + Shallow Network}} & \textbf{4.6} & 4.0 & 95\% & {\xmark} \\
& {GNN$_1$} & 5.2 & - & - & {\cmark} \\
\noalign{\hrule height 0.5pt}
\multirow{2}{*}{DCN} & \textbf{RaWaNet{\small{ + Shallow Network}}}  & \textbf{6.0} & 5.1 & 92\% & {\xmark} \\
& {GNN$_1$}  & {6.6} & - & - & {\cmark} \\
\noalign{\hrule height 0.5pt}
\multirow{3}{*}{BCF} & \textbf{RaWaNet{\small{ + Shallow Network}}}  & 0.38 & \textbf{0.36} & \textbf{89\%} & {\xmark} \\
& {GNN$_2$} & - & {0.48} & 82\% & {\cmark} \\
& {Zhao\:(QSAR)} \citeyearpar{zhao2008new} & - & - & 83\% & {\cmark} \\
\noalign{\hrule height 1pt}
\end{tabularx}
\end{threeparttable}%
\end{center}
\vskip -0.01in
\end{table*}

\subsection{MON, RON, DCN, BCF Property Prediction Tasks}
\citep{morris2019weisfeiler} introduce k-dimensional GNNs (k-GNNs) as a generalization to GNNs, which is capable of distinguishing non-isomorphism between graphs. They defined a local and global neighborhood for their introduced higher dimensional graph, which is based on higher order Weisfeiler-Leman graph isomorphisms. \citep{schweidtmann_graph_2020} designed a GNN for the prediction of MON, RON, DCN, three fuel-related molecular properties. They have implemented their networks using the 2-GNN from the k-GNN package to get higher-order graph structures as part of their message passing process. \citep{medina2021prediction} also implemented a message passing GNN using two layers of NNConv \citep{gilmer2017neural} followed by a GRU layer.

In our proposed network, as well as the GNNs that we compare our results to, we obtain an initial node embedding by applying a linear layer to the one-hot encoded node features. Here we use $X_1$, $X_2$ and $X_\gamma$ and take them to the dimension 100. We then apply a GraphNorm on outputs, followed by a ReLU function. After passing through a global add pooling we concatenate the feature vectors to the get the final embedding of the graph and feed it to a linear layer for prediction. The optimizer for all datasets is AdamW with a learning rate of $0.01$. The batch size for train/validation/test is 64/4/4 and the number of training epochs 200. We use StepLR scheduler for BCF dataset and ReduceLROnPlateau for the other three. All the shown experiments in Table \ref{tab:Reg2} are the average MAE over 40 runs with different random seeds for train/validation split. EL MAE and EL $R^2$ refer to the accuracy results from ensemble learning. Ensemble learning here means to use all the 40 trained models for prediction, and average over their results. As it is shown in Table \ref{tab:Reg2} we outperform all the tasks with a much faster and smaller network.

\section{Conclusion and Future Work}

In this paper we showed that a scaled input in combination with a shallow and simple architecture can show better or a on par performance with state of the art GNNs. For this, we introduced additional, more descriptive graph topologies along with their corresponding scaled node features. These are based on random walks of different lengths on the graph and their resulting stationary distributions as scaling of the node features. The resulting architectures are not only able to produce excellent performance but also only require orders of magnitude fewer parameters than most current GNNs rely on. Relying on simple networks we outperformed state of the art graph networks on HIV dataset and many more classification and regression tasks for molecular datasets.  

Our contribution is a data processing technique that enables us to encode more information through the additional inputs to the network, which grants us a simplified and efficient way of learning challenging datasets from various applications. We believe that this idea can be used within many different graph based architectures and unfold a new chapter of graph neural network architecture design.

\bibliographystyle{unsrtnat}
\bibliography{cite}  






\end{document}